\documentclass[conference]{IEEEtran}
\IEEEoverridecommandlockouts
\usepackage{cite}
\usepackage{amsmath,amssymb,amsfonts}
\usepackage{algorithmic}
\usepackage{graphicx}
\usepackage{textcomp}
\usepackage{xcolor}
\usepackage{balance}
\def\BibTeX{{\rm B\kern-.05em{\sc i\kern-.025em b}\kern-.08em
    T\kern-.1667em\lower.7ex\hbox{E}\kern-.125emX}}
\begin{document}

\title{3D Head-Position Prediction in First-Person View by Considering Head Pose for Human-Robot Eye Contact\\}

\makeatletter
\newcommand{\linebreakand}{%
  \end{@IEEEauthorhalign}
  \hfill\mbox{}\par
  \mbox{}\hfill\begin{@IEEEauthorhalign}
}
\makeatother

\author{\IEEEauthorblockN{Yuki Tamaru}
\IEEEauthorblockA{\textit{The University of Tokyo} \\
tamaru-yuki385@g.ecc.u-tokyo.ac.jp }
\and
\IEEEauthorblockN{Yasunori Ozaki}
\IEEEauthorblockA{\textit{CyberAgent, Inc.} \\
ozaki\_yasunori@cyberagent.co.jp }
\and
\IEEEauthorblockN{Yuki Okafuji}
\IEEEauthorblockA{\textit{Ritsumeikan University} \\
yokafuji@fc.ritsumei.ac.jp }
\linebreakand
\IEEEauthorblockN{Junya Nakanishi}
\IEEEauthorblockA{\textit{Osaka University} \\
nakanishi.junya@irl.sys.es.osaka-u.ac.jp}
\and
\IEEEauthorblockN{Yuichiro Yoshikawa}
\IEEEauthorblockA{\textit{Osaka University} \\
yoshikawa@irl.sys.es.osaka-u.ac.jp}
\and
\IEEEauthorblockN{Jun Baba}
\IEEEauthorblockA{\textit{CyberAgent, Inc.} \\
baba\_jun@cyberagent.co.jp}
}

\maketitle

\begin{abstract}
For a humanoid robot to make eye contact and initiate communication with a person, it is necessary to estimate the person's head position.
However, eye contact becomes difficult due to the mechanical delay of the robot when the person is moving. 
Owing to these issues, it is important to conduct a head-position prediction to mitigate the effect of the delay in the robot motion.
Based on the fact that humans turn their heads before changing direction while walking, we hypothesized that the accuracy of three-dimensional (3D) head-position prediction from a first-person view can be improved by considering the head pose.
We compared our method with a conventional Kalman filter-based approach, and found our method to be more accurate. The experiment results show that considering the head pose helps improve the accuracy of 3D head-position prediction.
\end{abstract}


\section{Introduction}

Robots have been increasingly used as substitutes to humans, including in commercial facilities.
In particular, humanoid robots can elicit attention by making eye contact to initiate communication with passersby \cite{admoni2017social, ono2000reading}.
When a humanoid robot makes eye contact, it needs to estimate the head pose of the passerby.
However, when the robot moves its gaze toward the head, it often fails to implement the intended action.
This occurs because the simple method of moving the gaze to the observed position of the subject's head when the subject is moving causes a mismatch between the gaze direction and the head position owing to the mechanical delay of the robot.
This makes it difficult to attract the attention of the subject, and can interfere with the communication.
To address this issue, a control method for robots that can predict human motion and direct their gaze to a predicted location on the head has been proposed \cite{8673175}.
The purpose of this method is to mitigate the effect of a recognition delay by predicting the head position. 
The effectiveness of this method has been verified in practice.
This research is mainly based on the assumption that the person the robot is talking to is in front of the robot.
Because the depth of the head position is assumed to be constant, it is difficult to respond to changes in depth.
In our study, we assumed that the subject having a conversation with the robot is walking in various directions.
Because the displacement of the subject's position is relatively large, the accuracy of the head-position prediction must be improved to mitigate the effect of any delay that occurs.
In our study, we focused on real-time location prediction of passersby from the viewpoint of the robot, i.e., a first-person view.
\begin{figure}[]
  \centering
  \includegraphics[keepaspectratio, scale=0.4]{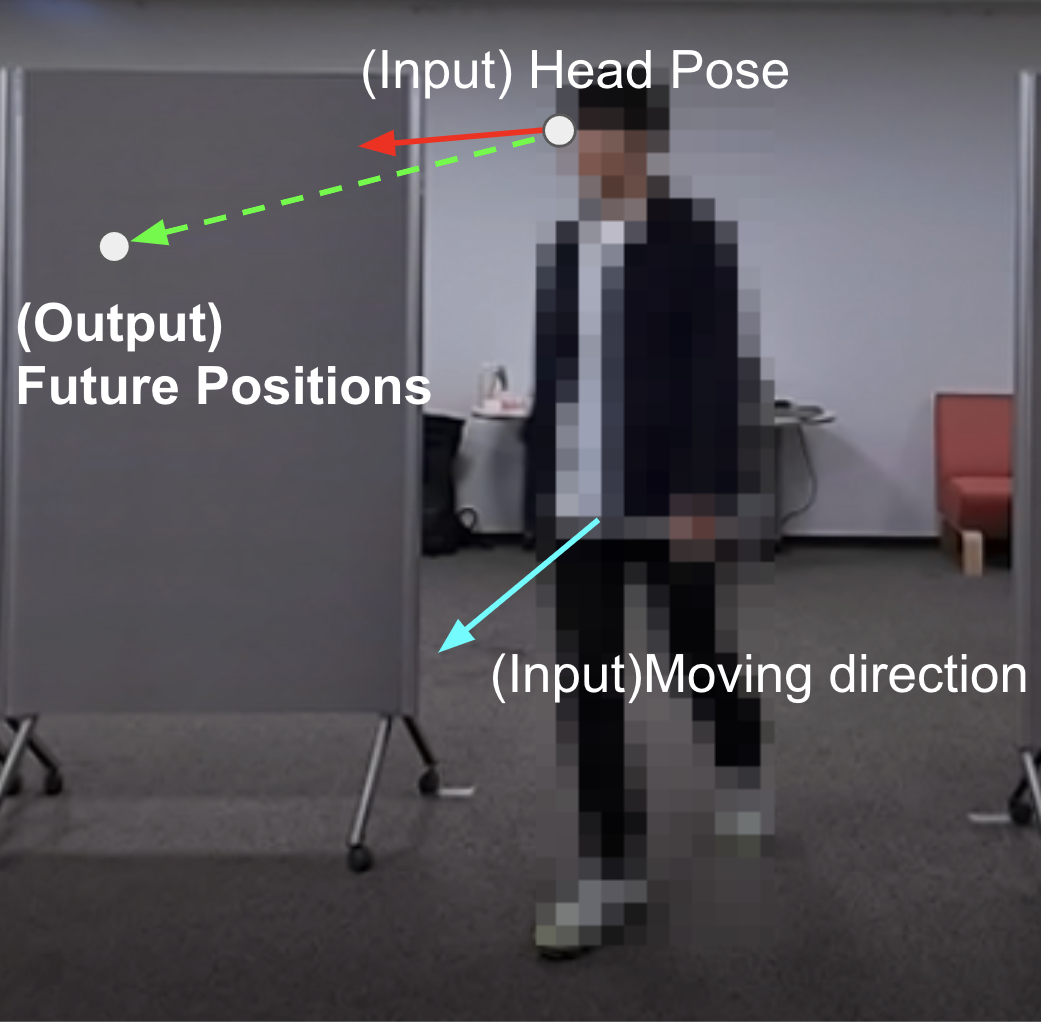}
  \caption{{\bf Head position prediction.} Given a first-person video of a certain target individual, our method predicts where the head of the target will be located in the future frames based on the head pose and moving direction.}
  \label{fig:Teaser}
\end{figure}

Humans typically turn their head toward the turning direction before turning the body as well.
If we can read the intentions of passersby regarding their moving direction,  it will be possible to predict the direction of movement with higher accuracy.
We believe it might be effective to consider the head direction when predicting the positions of the passersby.
However, we have not found any previous studies applying this hypothesis to the location of passersby from a first-person perspective.
In fact, passersby might walk in various directions before starting to communicate with a robot.

Through our study, we aimed to improve the accuracy of predicting the position of passers-by from a first-person perspective by considering their head pose.
We predict the 3D position of a passerby's head from the input of the 3D position of the RGB-D sensor.
To achieve this goal, we extract the position of each body part using an RGB-D sensor and predict the head position using a Kalman filter triggered by the head pose.

The contributions of our study are as follows:
\begin{itemize}
    \item We propose a method to predict the position of the head in 3D using RGB-D sensor information obtained from a first-person view. The prediction error using the proposed method was found to be less than that of a conventional approach.
\end{itemize}
\section{Related Work\label{sec:related}}
In this section, we introduce research related to human position prediction and the relationship between the direction of human movement and head pose.

\subsection{Human-Position Prediction}
Our research is classified as human-position prediction,
of which there are four types: 2D and 3D position predictions from a first-person view, and 2D and 3D position prediction from a bird's eye view, respectively.

A 2D position prediction from the first-person view was reported in a previous study \cite{Yagi_2018_CVPR}.
The goal is to predict a person's position in a future frame with a short first-person video continuously recorded by a wearable camera.
In their study, they also used human poses to predict future locations.
A prior study proposed a method for outputting the predicted rectangular region of a pedestrian using the current position, the vehicle's movement, and the onboard camera image as inputs \cite{bhattacharyya2018longterm}.

A study of 2D position prediction from a bird's eye view was also described \cite{Alahi_2016_CVPR}.
The authors presented a data-driven approach for learning these human-human interactions to predict future trajectories.
This makes it possible to predict paths and avoid collisions between people.
In another study \cite{hasan2018mxlstm}, the human trajectory, head direction, and interaction information are the inputs to the LSTM to achieve a path prediction that avoids collisions with other objects within the viewing angle.
In addition, by arbitrarily changing the gaze information, it is possible to predict the path toward any direction.
In particular, a prior work considered the orientation of the human head, and the effectiveness of these methods in practice was verified \cite{hasan2018mxlstm}.
Therefore, we hypothesize in our study that there is a certain effect when considering the head pose in a 3D position prediction.

A 3D position prediction through a bird's eye view was reported in \cite{6636027}.
In this study, a method for tracking the position, orientation, and height of people in public spaces is proposed.
In this study, a 3D distance sensor was used to obtain the angle of the body, and a particle filter \cite{doucet2000sequential} was used for tracking.

Our study can be regarded as a 3D position prediction from a first-person view.
However, we were unable to find any previous studies on a 3D position prediction from a first -person perspective, thereby indicating the high novelty of this research.

\subsection{Relationship between direction of human movement and head pose}
The motivation for considering the head pose is introduced in this section.
We know from a prior study \cite{hasan2018mxlstm} that there is a correlation between the direction of movement and the head pose.
This study demonstrated that the abovementioned correlation is statistically significant.
This suggests that it is useful to consider the head pose while estimating the direction of movement.
It was also reported by some previous studies \cite{grasso1998eye, courtine2003human, lopez2019walking} that the change in head orientation precedes the change in direction of movement.
In some studies \cite{courtine2003human, lopez2019walking}, head yaw has been observed to be the most reliable kinematic variable from the upper body, which predicts walking turns of approximately 200 ms.
The delay time of the robot's motion is approximately 500 ms, suggesting that it may be possible to predict the robot's motion by considering its head pose.
Based on these facts, we hypothesized that the introduction of head pose into a real-time 3D head position prediction will improve the accuracy, and we tested this hypothesis.
\section{Implementation}
In this section, we describe the head position recognition and the head position prediction methods allowing a robot to successfully look at a human head.
The pipeline of our proposed prediction method is shown in Fig.~\ref{fig:pipeline}.
\begin{figure*}[ht]
    \includegraphics[width=\textwidth]{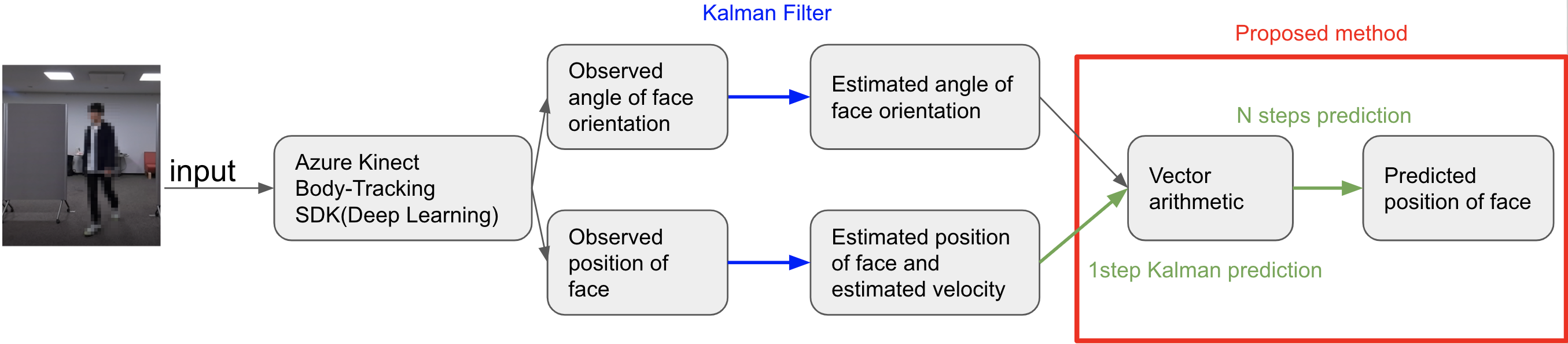}
    \caption{Implemented pipeline. The input is an RGB-D image. The output is the predicted position of the head.}
    \label{fig:pipeline}
\end{figure*}
The input is an RGB-D image captured by an Azure Kinect \cite{azurekinect}.
The observed angle of the head pose and observed position of the heads are extracted using an Azure Kinect Body-Tracking SDK \cite{azbt-sdk}.
Each input was then subjected to a Kalman filter \cite{kalman1960new} to calculate the estimated angle of the head pose, estimated position of the head, and estimated velocity.
Using the values obtained, we apply a Kalman prediction after 1 step considering the head pose, and use it to calculate the predicted position of the head after 500 ms.
In the abovementioned pipeline, head position recognition and head position prediction are conducted.
These are described in the following sections.

\subsection{Head-Position Recognition}
Head position recognition was achieved using an Azure Kinect Body-Tracking SDK, which can acquire and track human skeletal information in images using Azure Kinect, an RGB-D camera.
We used this Azure Kinect Body-Tracking SDK in our study because, as mentioned in Section \ref{sec:related}, we were unable to find any dataset for first-person 3D prediction, and it was difficult to build and train such a model.
We used the coordinates and quaternions of the nose as the position and orientation of the head.
For the experiment, we defined the head pose as the relative angle of the nose orientation to the waist orientation.

\subsection{Head Position Prediction}
In this section, we describe a head position prediction method.
To conduct experiments focusing on a real-time prediction, we investigated various prediction methods based on the use of a Kalman filter.

\subsubsection{Proposed Method}
The proposed method is based on a Kalman filter; however, the method of calculating the predicted state estimate for N-step prediction is different from a conventional Kalman filter.
To predict the 3D position of the head, the state variable $\boldsymbol{x}$ was set to six dimensions, including 3D position and velocity.
The predicted state estimate $\hat{\boldsymbol{x}}_{t+N|t}^{p}$ after N steps at time $t$ was obtained through the procedure denoted in the equation below.
Note that $\boldsymbol{R}$ is a 3-dimensional rotation matrix corresponding to the angle of head pose.
\begin{equation}\label{eq:t+1}
    \hat{\boldsymbol{x}}_{t+1|t}=\boldsymbol{F}\hat{\boldsymbol{x}}_{t|t}
\end{equation}
\begin{equation}\label{eq:dk}
    \boldsymbol{d}^{kalman}=\hat{\boldsymbol{x}}_{t+1|t}-\hat{\boldsymbol{x}}_{t|t}
\end{equation}
\begin{equation}\label{eq:df}
    \boldsymbol{d}^{head}=\boldsymbol{R}\boldsymbol{d}^{kalman}
\end{equation}
\begin{equation}\label{eq:dp}
    \boldsymbol{d}^{p} = (1-\boldsymbol{w})\boldsymbol{d}^{kalman}+\boldsymbol{w}\boldsymbol{d}^{head}
\end{equation}
\begin{equation}\label{eq:t+n}
    \hat{\boldsymbol{x}}_{t+N|t}^{p}=\hat{\boldsymbol{x}}_{t|t}+N\boldsymbol{d}^{p}
\end{equation}
First, we calculated the predicted state estimate of $t+1$ one step later by using the Kalman filter prediction step in Equation~(\ref{eq:t+1}).
Next, we calculated $\boldsymbol{d}^{kalman}$, which is the difference between the predicted state estimate one step later at time $t$ and the predicted state estimate at time $t$ in Equation~(\ref{eq:dk}).
Next, we calculated $\boldsymbol{d}^{head}$, which is a rotation of $\boldsymbol{d}^{kalman}$ based on the rotation matrix $\boldsymbol{R}$ in Equation~(\ref{eq:df}).
$\boldsymbol{w}$($\in[0,1]$) is the weight, and $\boldsymbol{d}^{p}$ is determined by $\boldsymbol{w}$.
Finally, we calculated the state estimate $\hat{\boldsymbol{x}}_{t+N|t}^{p}$ after N steps using Equation~(\ref{eq:t+n}).
Through this approach, we obtained the state estimates after N steps at each time.
Fig.~\ref{fig:pm} briefly illustrates the above procedure.
\begin{figure}[]
    \includegraphics[width=\linewidth]{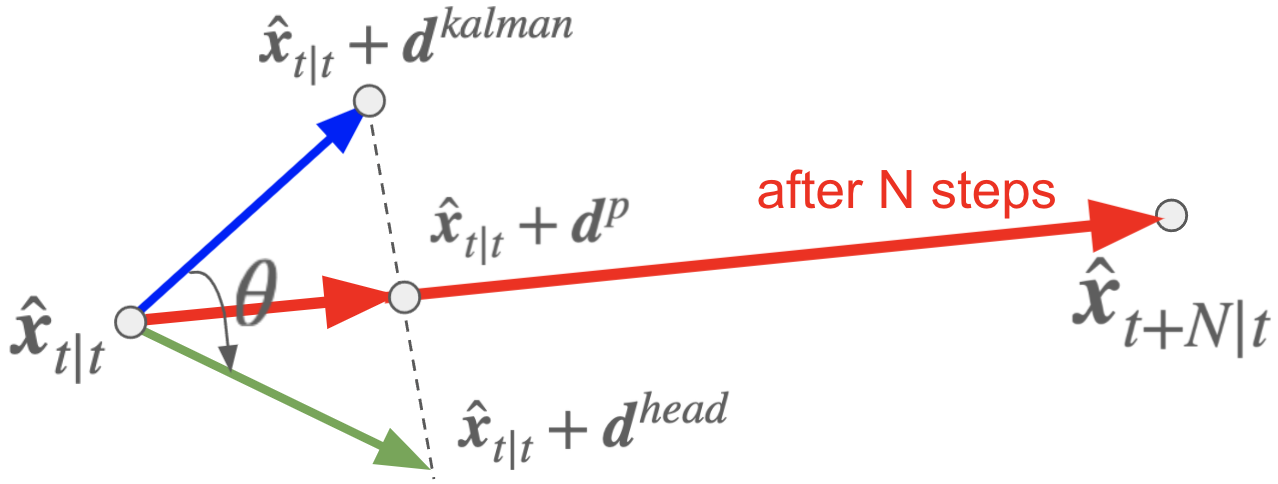}
    \caption{Visualized example of the proposed method. $\theta$ is the angle of the head pose and is used for the rotation matrix $\boldsymbol{R}$.}
    \label{fig:pm}
\end{figure}
\section{Data Acquisition}

\subsection{Experiment}
In this section, we describe the procedure of the data acquisition experiment.
The purpose of this experiment was to obtain a dataset of RGB-D videos of the subject while walking.
The experiment environment is as shown in Fig.~\ref{fig:env}.
\begin{figure}[hbtp]
    \includegraphics[width=\linewidth]{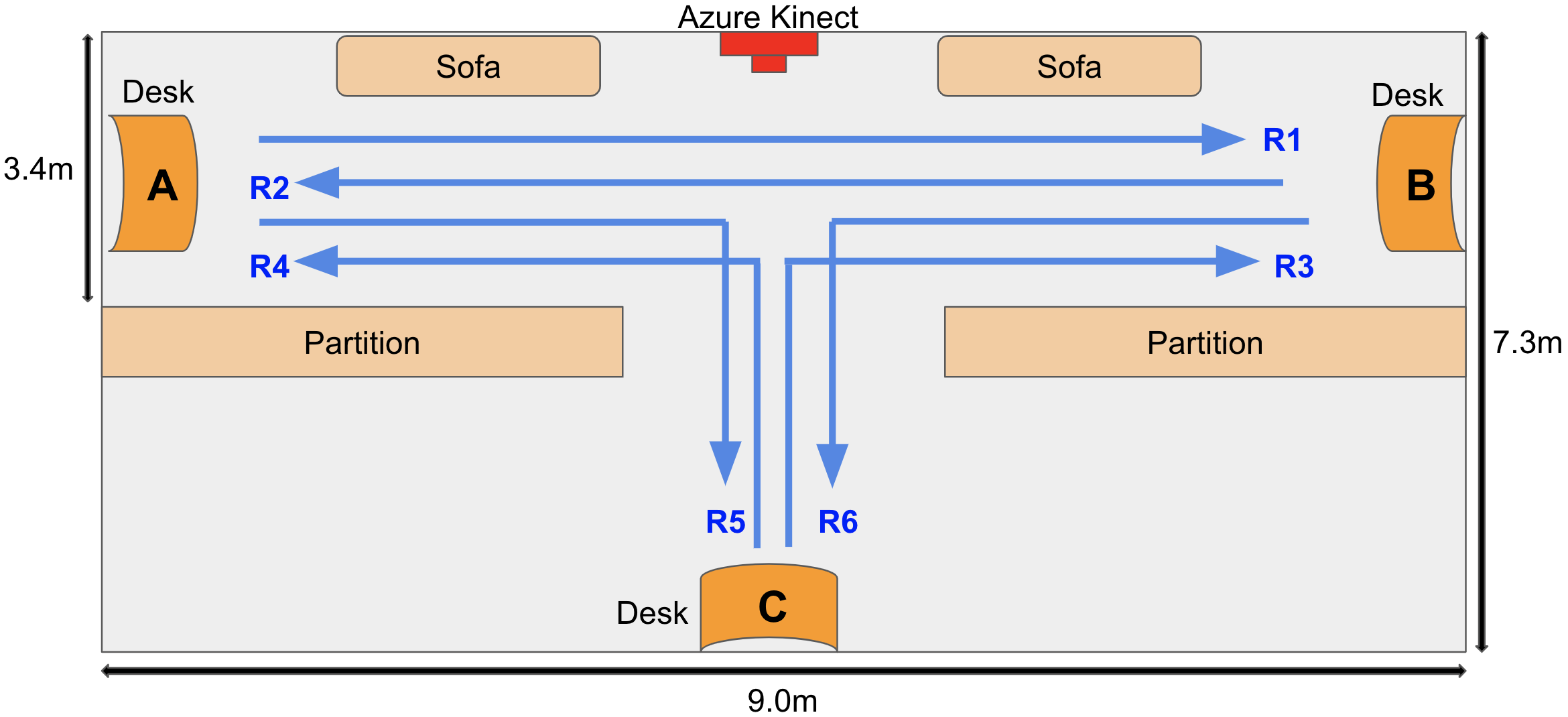}
    \caption{Overhead view of the experiment environment.}
    \label{fig:env}
\end{figure}
In the experiment room, we set up three landmark desks and an Azure Kinect.
The rooms were separated by partitions, and thus each landmark is not directly visible from the other landmarks.
The size of the room is 7.3 m × 9.0 m square, as shown in Fig.~\ref{fig:env}.
Data acquisition experiments were conducted on a total of 14 subjects.
The walking routes of R1-6 for the subjects are shown in Fig.~\ref{fig:env}.
For landmarks, desks were placed at the beginning and end of the routes.
In this experiment, the depth mode of the Azure Kinect was set to NFOV 2 × 2 Binned (SW).
For NFOV 2 × 2 Binned (SW), the operating range is from 0.5 to 5.46 m.
The videos were taken at 30 fps.

\subsection{Results}
In this section, we describe the procedure of the data acquisition experiment.
We obtained a total of 84 videos, each approximately 5 s long.
\section{Hypothesis Validation\label{sec:hv}}

\subsection{Experiment}
In this section, we describe the procedure used in the hypothesis validation experiment.
The purpose of the evaluation experiment was to compare the accuracy of our method and conventional Kalman filter prediction (hereafter referred to as the baseline method) on the acquired video dataset.
We used the data within the operating range.
The horizontal angle of the waist was defined as the moving direction, and the difference between the horizontal angle of the waist and the horizontal angle of the head was defined as the angle of the head pose.
The head and moving directions were input into the proposed and baseline methods.
The predicted values after 0.5 s, or 15 steps in our setting, were calculated for each frame.
The values obtained by the Azure Kinect body-tracking SDK are referred to herein as the observed values.
For the routes R1-6 set in the data acquisition experiment, we grouped the left and right target routes. 
We label the group of R1 and R2 as R12, R3 and R4 as R34, and R5 and R6 as R56.
Each group has 28 videos.
The calculated positions are mapped to the symbols as follows:
\begin{itemize}
    \item[(a)] The estimated position with a Kalman filter applied to the observed position of head.
    \item[(b)] Predicted position of the head using the baseline method.
    \item[(c)] Predicted position of the head using the proposed method.
\end{itemize}
The proposed method was cross-validated with a leave one subject out approach.
The validation method is as follows:
\begin{enumerate}
    \item Using the training data, we conducted hyperparameter tuning on $\boldsymbol{w}$ with Optuna\cite{akiba2019optuna}. The objective function was the sum of the errors for all videos in (c) and (a), and $\boldsymbol{w}$ was tuned to minimize such errors.
    \item For each group, we compared the error between (c) and (a) and the error between (b) and (a) on the test data. We used one-tailed tests of the Wilcoxon signed-rank tests to test for significant differences. We set the significance level $\alpha$ to 0.05.
\end{enumerate}
The error was calculated using the Euclidean distance.

The hypotheses tested in this experiment are as follows.
\begin{itemize}
    \item For the turning route (R3-6), the accuracy of the proposed method is better than that of the baseline method because the direction in which the head moves is believed to precede the travel direction.
    \item For the straight route (R1-2), the results of the baseline and proposed methods are almost identical because the direction in which the head moves is believed to always match the travel direction.
\end{itemize}

\subsection{Results}
\begin{figure}[]
    \centering
    \includegraphics[scale=0.2]{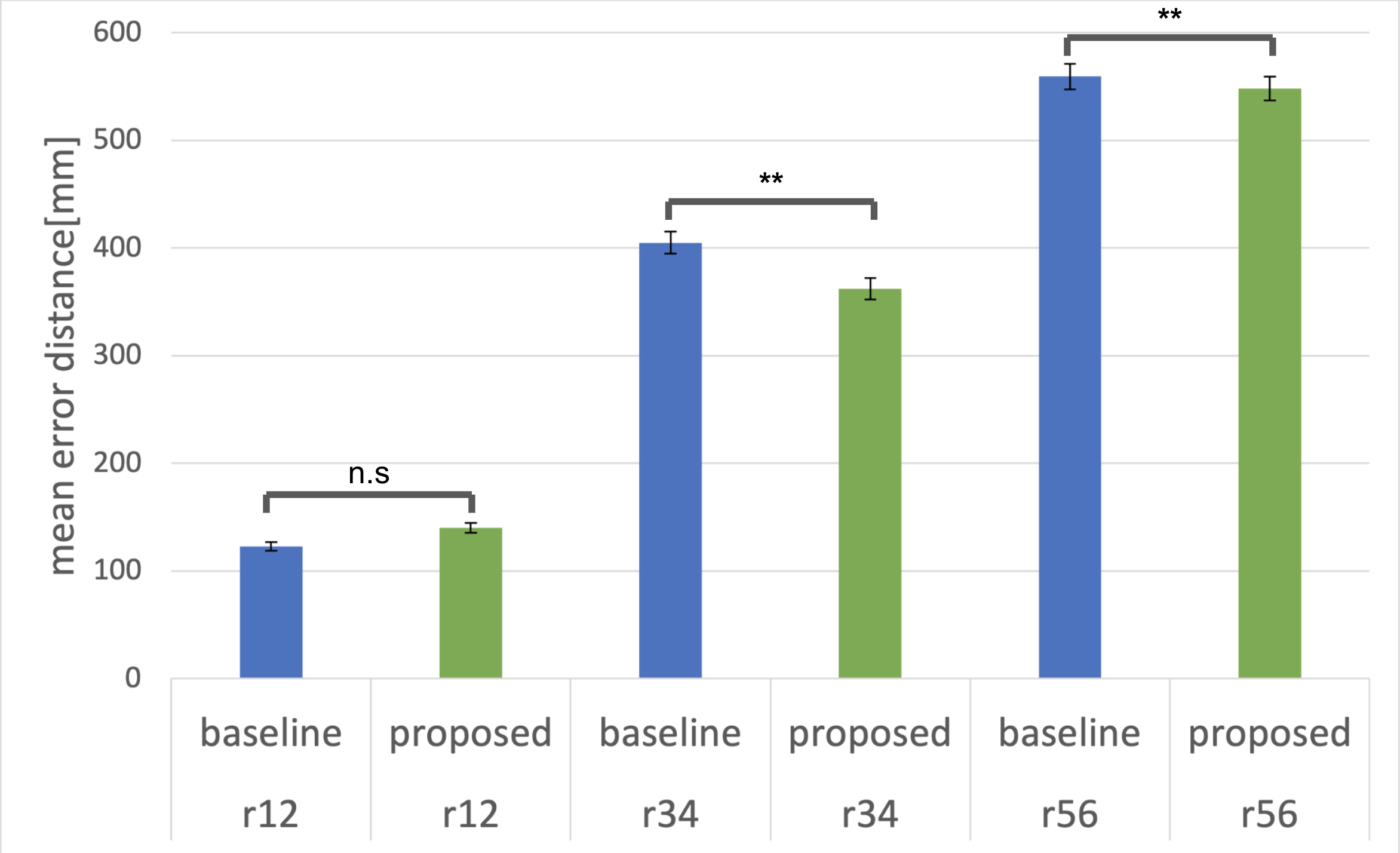}
    \caption{Mean error distance of the validation experiment under each method.}
    \label{fig:result}
\end{figure}
The results of a cross validation, in which $\boldsymbol{w}$ was conducted hyperparameter tuning, showed that the mean of $\boldsymbol{w}$ was 7.51, and the standard deviation was 0.36.
Fig.~\ref{fig:result} shows the mean error distance for each method.
For R12, the mean error was 122 mm for the baseline method and 139 mm for the proposed method, with that of the baseline method being lower.
The results of the Wilcoxon signed-rank test showed that $p=1.0 >0.05$.
The statistic was 579,000.
Therefore, no significant differences were observed.  

For R34, the mean error was 404 mm for the baseline method and 362 mm for the proposed method, when applying the error observed with the proposed method being lower.
The results of the Wilcoxon signed-rank test showed that $p=6.09\times10^{-93}<0.05$.
The statistic was 2,849,081.
Therefore, a significant difference was observed.

For R56, the mean error was 559 mm for the baseline method and 548 mm for the proposed method, the error when applying the proposed method being lower.
The results of the Wilcoxon signed-rank test showed that $p=1.62\times10^{-3}<0.05$.
The statistic was 1,990,900.
Therefore, a significant difference was observed.

\section{Discussion}
The purpose of this experiment was to improve the accuracy of a 3D head position prediction of passersby by considering the head pose.
In this experiment, we tested the hypothesis as mentioned in Section~\ref{sec:hv}.
The results indicate that they are in line with the hypothesis.

For this method to be effectively applied, it is desirable to improve the accuracy of the method for head pose recognition from a first-person view.

\section{Conclusions}
In this study, we proposed a prediction method to improve the accuracy of 3D head-position prediction of passersby when considering the head pose. From the experiment results, it is evident that taking head pose into account is effective in predicting the position of a person in 3D in real time.
In a future study, the proposed method can be combined with existing head pose estimation methods such as those described in \cite{ruiz2018finegrained, yang2019fsa, hsu2018quatnet}.

\bibliographystyle{./IEEEtran} 
\balance
\bibliography{./IEEEfull,./IEEEexample}

\vspace{12pt}

\end{document}